# Applied Federated Learning: Architectural Design for Robust and Efficient Learning in Privacy Aware Settings

Branislav Stojkovic[1], Jonathan Woodbridge[1], Zhihan Fang[1], Jerry Cai[1], Andrey Petrov[1], Sathya Iyer[1], Daoyu Huang[1], Patrick Yau[1], Arvind Sastha Kumar[1], Hitesh Jawa[1], Anamita Guha[1]

## Abstract

The classical machine learning paradigm requires the aggregation of user data in a central location where machine learning practitioners can pre-process data, calculate features, tune models and evaluate performance. The advantage of this approach includes leveraging high performance hardware (such as GPUs) and the ability of machine learning practitioners to do in depth data analysis to improve model performance. However, these advantages may come at a cost to data privacy. User data is collected, aggregated and stored on centralized servers for model development. Centralization of data poses risks, including a heightened risk of internal and external security incidents as well as accidental data misuse. Federated learning with differential privacy is designed to avoid the server-side centralization pitfall by bringing the ML learning step to users' devices. Learning is done in a federated manner where each mobile device runs a training loop on a local copy of a model. Updates from on-device models are sent to the server via encrypted communication and through differential privacy to improve the global model. In this paradigm, users' personal data remains on their devices. Surprisingly, model training in this manner comes at a fairly minimal degradation in model performance. However, federated learning comes with many other challenges due to its distributed nature, heterogeneous compute environments and lack of data visibility. This paper explores those challenges and outlines an architectural design solution we are exploring and testing to productionize federated learning at Meta scale.

[1] Meta

# Introduction

Federated learning (FL) has witnessed a remarkable growth in popularity in recent years with applications ranging from consumer devices to healthcare and fintech [1,2,3]. Federated learning enables data to stay on users' devices by training models in a completely distributed way. Once trained, inferences can occur completely on mobile devices with no personal data being sent to the backend servers. Back-propagation occurs on devices and only model updates are sent to the server to be aggregated. For example, applications running on mobile devices, such as Facebook and Instagram, can continue to deliver the same seamless and intelligent user experience without the data ever leaving the device.

Differential privacy adds another layer of privacy by adding noise to model updates. As an example, authors in [6] propose an algorithm for stochastic gradient descent (SGD) that applies clipping and noise to gradients after each step of gradient descent. Random noise reduces the ability to reverse training examples while gradient clipping minimizes the contribution of each training sample (i.e., minimizes the ability to memorize users' data) [7].

Combining federated learning with differential privacy results in three main benefits:
1. Data remains on user devices;
2. Aggregation of client data is performed within a trusted environment that is not accessible by corporate servers. Only aggregate information reaches corporate servers for product use;
3. Model are unlikely to contain memorized personal data

However, the privacy benefits of federated learning with differential privacy also presents challenges. Without visibility of data, ML practitioners can no longer rely on conventional exploratory data analysis, model tuning and debugging techniques thereby significantly slowing down model development.

Meta has taken the initiative to increase the adoption of federated learning with differential privacy for both research-exploratory purposes and potential production deployment. This effort has uncovered many new challenges that are unique to federated learning with differential privacy. Some notable examples include **i) Label balancing, feature normalization and metrics calculation** due to lack of data visibility; **ii) Slower mobile release cycles** as compared to backend release cycles; **iii) Slower training** due to federation of training to mobile devices; **iv) De-identified system logging** is required to promote data privacy.

This paper presents an architecture that addresses the aforementioned challenges and has the potential to scale inferences to billions of devices. The focus of this work is creating binary classifiers for slowly moving datasets that were historically computed on the server side. This architecture enables model training to combine server-side user data with device-side-only user data to deliver inferences. This enables device-side-only user data to be generated and stay on users' devices. This architecture is a combination of infrastructure across mobile devices, trusted execution environments and conventional backend servers. Validation of this architecture is performed on an in-house federated learning library that is compatible with Meta's family of apps (such as Facebook and Instagram) and has the potential to scale training to millions of devices. This approach is compared to conventional server trained models and demonstrates minimal degradation of model performance without transgressing constraints of limited on-device compute, storage and power resources.

In summary, this paper has two main contributions:
1. Highlight the challenges of productionizing federated machine learning; and
2. Propose a federated learning architecture to solve those challenges.

This paper is structured as follows. In Section 2, we explain in detail the challenges we face in our endeavor. Section 3 is dedicated to overview of the architecture of our system and advantages it offers for training and deploying models in production. Section 4 goes into our particular techniques we use to overcome the challenges. In Section 5, we talk about results we observed in deploying the federated trained model with respect to expected versus actual metrics drop-off from server side model.

## Challenges of Federated Learning

Federated learning with differential privacy allows enhanced user experiences driven through ML while increasing user control over data. However, not being able to access the original data on the central server introduces some fundamental challenges that significantly slow down the model development process. ML practitioners are forced to rethink some basic tasks that were previously considered solved. In this paper we highlight six challenges we faced while developing federated learning at Meta and describe methods to mitigate them:
1. Label balancing
2. Slow release cycles
3. Low device participation rate
4. Privacy preserving system logging
5. Model metric calculation
6. Feature normalization

## Label Balancing

In many real-world applications, the corresponding classification task is of an imbalanced nature. Using imbalanced labeled data in training can lead to a significant loss of performance in most classifier learning algorithms, as they expect a balanced class distribution.

In federated learning, label balancing is difficult as there is no information sharing between devices other than the aggregation of weights at the server. In addition, back-propagation is computed on each device meaning the server is blind to any labels that can be used to do global label balancing. When dealing with a class distribution that is long-tailed, label imbalance becomes a major barrier as training a classifier in a dedicated window of time becomes prohibitively harder due to slow loss convergence.

## Slow Release Cycles

Updating a mobile codebase requires a "release" which usually occurs on iteration cycles that are orders of magnitude slower than server iterations. Companies may release a couple times a day, week, month or longer. This means that a simple mobile update (such as adding a feature) could take hours, weeks, or months before a large enough user base can be used to train a model. Additionally, some users rarely update their apps which made keeping the code bugfree extremely difficult and caused some historical inefficiencies. We invested a significant amount of effort to make critical functionality of our code independent of an app update event.

## Low Device Participation Rate

Mobile devices are highly resource constrained. Without care, training and inference can negatively impact device battery level, processing, storage and network bandwidth utilization. These regressions will result in a poor user experience and may result in users deleting the app from their devices. Our goal is to deliver improved privacy-aware user experiences with no noticeable regressions to our users. For this reason, not all eligible devices participate in training of a federated learning model. There is a set of carefully crafted heuristics implemented within the native app that serve as a safeguard against potential regressions and determine eventual device participation.

## Privacy Preserving System Logging

Contemporary native app development and maintenance heavily rely upon monitoring to understand how system components interact and behave in production. The data provided by native app logs are critical to ensure high quality user experience by allowing for efficient detection and diagnosis of undesired behavior. Logging and debugging are challenging due to the privacy sensitive nature of federated learning.  When logging, we cannot centralize any identifying information as this would undo the original purpose of deploying federated learning with differential privacy.  However, this adds challenges to debugging both our code implementation as well as any ML model inefficiencies. In addition, it adds a critical point of failure where a developer could accidently log user information when trying to debug.

## Model Metric Calculation

Suboptimal model performance quality usually leads to inferior user experience upon deployment in production. Model evaluation metrics play a critical role in achieving the optimal model performance during the training process. Calculating  these metrics is an additional challenge.  A naive solution for measurement without adding bias would be to upload model predictions alongside raw feature or label values from a randomly sampled population of users (or devices) participating in evaluation. This would enable calculating metrics such as precision/recall, ROC AUC, etc.  In addition, the features and metadata would enable debugging to understand potential bias and other shortcomings of the model. However, sending any information back to the server that could be later associated with the original user would undo the intent of using federated learning with differential privacy.

## Feature Normalization

The architecture presented in this paper is limited to neural networks which largely benefit from feature normalization.  Without normalization, neural networks are slow to converge (or may not converge at all).  In server based ML, normalization is a fairly trivial operation with normalization factors determined on the training set.  In the federated space, there is no information sharing between nodes except for the aggregation of model weights at a central server.  This requires additional functionality built within the architecture to learn normalization factors.

# Architecture

The fundamental driving principle behind our infrastructure architecture design decisions was to improve developer efficiency. While there are a lot of results on enabling successful and efficient model training [5], not much attention is paid to auxiliary core infrastructure components that are needed for speedy tuning and scalable deployment at inference time. For this reason, we chose to focus on the overall architectural design and integration. Here, we outline the major components of the system. While outside the focus of this paper, the reader should assume all communications are properly encrypted.

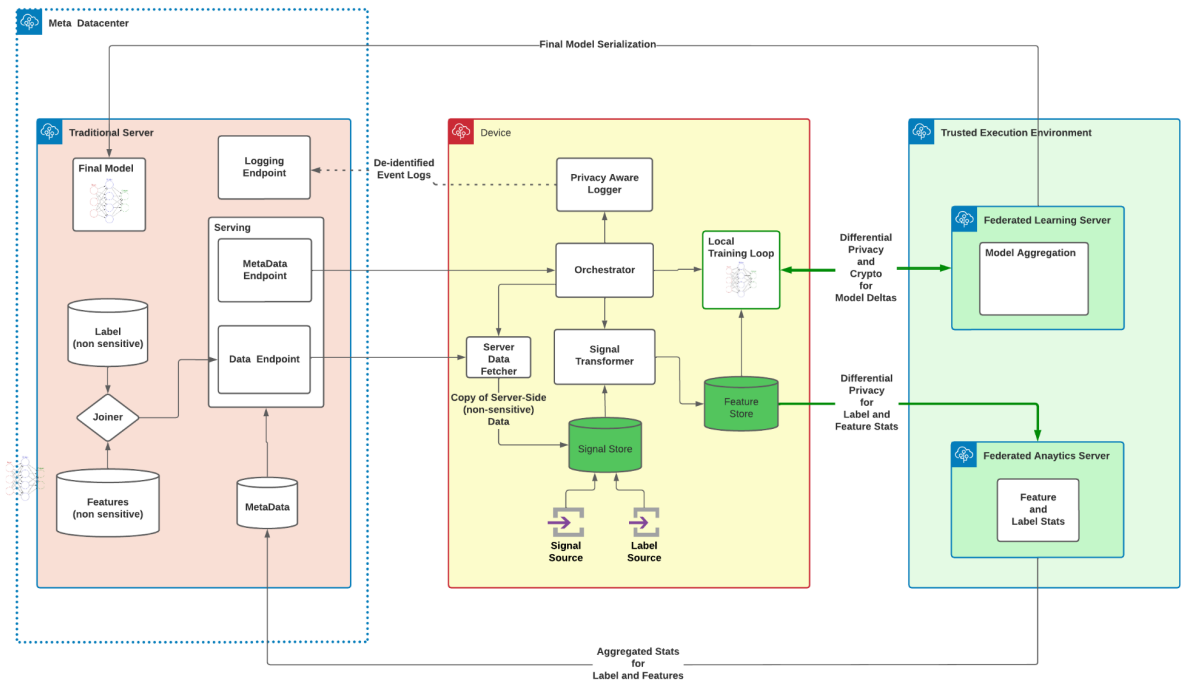

Figure 1: DCP Architecture Overview

**Federated Learning Server (Trusted Execution Environment (TEE))** Server running an implementation of synchronous secure aggregation protocol for Federated Learning with Differential Privacy. Hardware-based support for encryption and computation ensures that unauthorized entities, including traditional Meta servers, cannot view or update the data while it is in transit or in use within the TEE. In this manner, we can consider this setup an extension of the user's device as it offers comparable control and protection of data. The server sends a snapshot of the global model to each of the eligible devices. Devices use local samples to execute forward pass and subsequently backpropagation to update the weights. The server waits for the participating devices to report local updates to the model. Once a desired number of updates has been received, the server aggregates them using weighted averaging. The process continues until enough devices report the updates at which point the round is marked as completed. Training is completed when desired accuracy is achieved, usually after several rounds.

**Federated Analytics Server (Trusted Execution Environment)** Server supporting Differential Privacy computation at scale. Scale is important in computing basic statistics about features and label distribution as these estimates are often obtained on orders of magnitude larger population size than the actual on-device model training one. In our application we use a protocol for computing means and percentiles based on a manipulation of individual bit values [4]. This easy to implement solution fits our scalability needs while providing the similar privacy considerations as other state-of-the-art methodologies.

**Model** type and architecture are predefined ahead of the deployment. In our implementation we rely solely upon dense features to even further reduce the chance of memorizing individual data entries during training. Initial hyperparameter tuning is performed on the server side. Server-side only available data are used for this purpose. During this phase, the neural network width, number of hidden layers and learning rate are determined. The model itself is written in PyTorch scriptable format.

**Features** (independent variables) usually have three potential origins: (1) server side, (2) device side or (3) both. Examples for features with origin in (1) may include engagement information such as historical interactions with content. Features with origin (2) are used consistent with user permissions and confined to devices to provide an additional layer of privacy for our users. (e.g. metadata about the device's phone number). Examples of features in (3) include server side information that is also available on device. The primary motivation of such features is to increase resolution and/or lower latency by using the signal directly from the device (e.g. metadata about feed scrolling speed and pause frequency). In scenario (3), whenever available we overwrite server side values with those computed on device.

**Label** is a dependent variable that we want our model to predict. For the purpose of this paper we limit ourselves to binary classification problems (e.g. labels can have two potential values). Examples of server-side labels are click or conversion events given an impression or a label generated by a human rater. Examples of on-device-only labels may include user's real time interactions with product surfaces.

**Joiner** This server-side process performs an action of assigning a label to a set of features. In a classical setup, after this step is completed, the pair becomes an input for model training. In our scenario, we often augment the feature set on a device with some additional signals and sometimes even update the label prior to the training. On device, this augmentation process is handled by *Signal Transformer*.

**Orchestrator** is the main component that coordinates device processes outside of local training. Orchestrator coordinates across several use cases. Here are the most critical tasks that are performed by Orchestrator: (1) scheduling, (2) running user/device eligibility checks, (2) server-to-device data flow initialization (3) control of submission of a sample for training and (4) logging and perf metric computation.

**Signal Transformer** is the core ML infra component on the device. It performs several critical tasks that include: local signal transformation into feature, local feature normalization, server side feature injections and local value overrides. Signal transformer is implemented in Pytorch and can be dynamically pushed to devices upon an update.

**Local Device Storage** Encrypted storage on device with a dedicated purpose to support federated learning and inference. This storage is separated from other storage on the device. It is a general solution that could support other (potentially non-ML related) cases.

**Server-side data/metadata serving** Endpoints that provide metadata to support running training and inference for several use cases. Examples of metadata include eligibility criteria (to be verified on device), model version, purpose associated with data, etc.

## Addressing Challenges through Proposed Architecture

It's helpful to view the lifecycle of a model trained using federated learning in order to understand how our architecture solves the six highlighted challenges in Federated Learning. Figure 2 shows the lifecycle in respect to the trusted compute components of our architecture. We consider both the Trusted Execution Environments and mobile devices as trusted as these environments (and respective data) are inaccessible by Meta internal servers and employees.

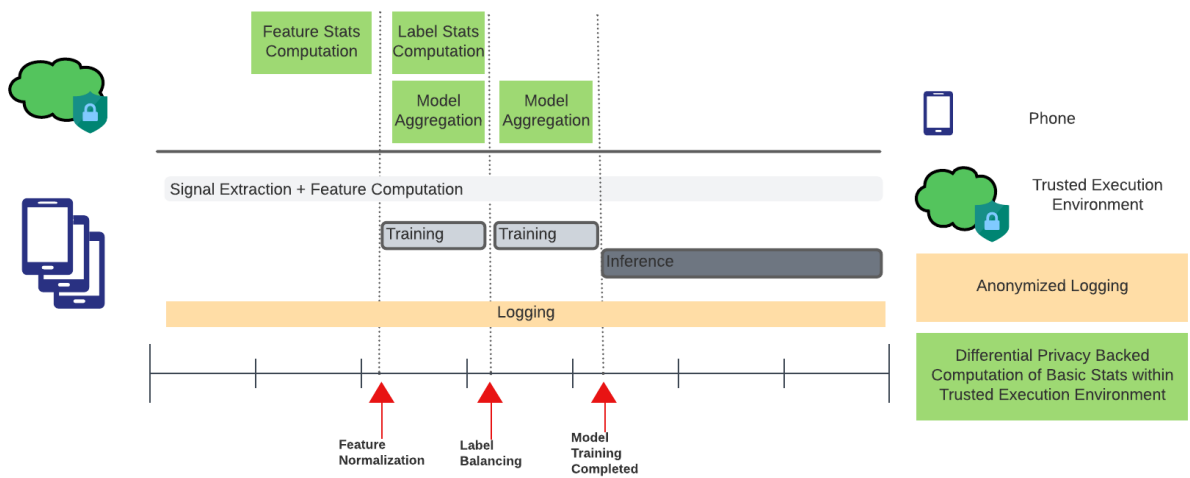

Figure 2: There are three functions highlighted in this timeline that occur within the Trusted Execution environment including computation of feature statistics, computation of label statistics and model aggregation.

## Trusted Execution Environment

**Computation of feature statistics** is how we solve feature normalization. This process is often referred to as Federated Analytics. Feature statistics are computed over a random sampling of our mobile device universe. Statistics are computed within the trusted environments such that only the aggregated data that is not unique to a particular user is eventually sent to company servers. To promote privacy the devices used to compute these statistics are selected independently of training meaning the same devices may not be used for both statistics and training.

**Computation of label statistics** is how we solve for label imbalance. During this process, we treat the label as yet another feature. Upon extraction, we store labels alongside associated feature sets within the feature store. To compute aggregated and noisy statistics for the labels we rely upon the same process we used for feature statistics computation. Upon export from a trusted environment, aggregated label statistics are exported and persisted in the metadata store on traditional servers. During training, the drop off rate is adjusted based on the most recent values in the metadata store. On device this value is used by Orchestrator to control sample submission.

**Model aggregation** is where model updates are aggregated over many devices. We can also add an optimization to help the model converge faster during this step. We have two choices on where to apply differential privacy: 1.) on device 2.) on the trusted execution environment. In case 1, noise is added to the model updates before leaving the device. In case 2, noise is added after the aggregation, but before applying to the global model update. In either case, the global model is only updated with weights after noise is added. The advantage to adding noise at the trusted execution environment is faster convergence and more accurate models. However, models trained using this optimization are still far slower than training in the cloud in centralized servers.

## Mobile Devices

There are four main functions highlighted in Figure 2 that include signal extraction + feature extraction, training, inference and logging. For simplicity, we are collapsing all stages of federated learning on mobile devices into a single timeline. In reality, each device may not participate in all stages (e.g., a device may participate in training, but not inference).

**Signal extraction + feature extraction + label computation** occurs first in our timeline. We first need to collect features before we can compute statistics, train or perform inference. In some cases, features may be computed over a time window (such as time spent in an application). For this reason, we often have a warm up period to compute features before we can begin training. Feature extraction is also an area where we can optimize slow release cycles. Instead of computing features in native mobile code, we use torch script. This allows us to download feature computation directly to the device without updating the application. This reduces the dev cycle of features from weeks to hours.

**Training** in federated learning is much slower than more traditional centralized training performed on servers. In addition, a high amount of network overhead can occur when passing models and model updates between mobile devices and servers. One optimization is to deploy an asynchronous federate learning architecture [5] which can decrease training times by 5x and reduce network overhead by 8x.

**Metric calculation** happens during the training process when we set aside a dedicated subset of the user population to compute relevant model performance attributes. User data that participates in computation of evaluation metric stays on the device. The actual metrics results derived from this data have statistical noise added to them and are being sent to our Federated Learning Server via encrypted channels. Later on, the results are exported to traditional servers for consumption (e.g. dashboarding and alerting) without any user identity being sent with them.

**Inference** is handled independently of the training flow. However, they are both built on top of the Feature Store as a shared foundation that ensures computational signal processing equivalence. Global model binaries are requested and fetched from server-side using traditional infrastructure. Models are stored locally and loaded into memory during the inference phase. For the inference on devices we rely upon Pytorch Mobile modeling libraries which utilize the TorchScript engine. These libraries provide APIs that cover common preprocessing and integration tasks needed for incorporating models in mobile applications (e.g. efficient model quantization, model binary downloading, loading, computing prediction).

**Logging** plays a critical role in supporting development on devices. In a complex system built of several components and product integration the need arises for funnel logging. For this purpose we divide the dataflow into phases and each phase can be further divided into steps. Logs from all successful and failed steps from a current phase should add up to the count of successful steps from the previous phase. By understanding where the drop off is happening we are able to effectively identify the issues or opportunities in our design. For the purpose of deduping logging events across different use cases ephemeral, randomly generated ids are assigned to each session. Session is defined as a time interval during which the user engages with a single product surface. These session level ids cannot be traced back to the original user.

For our system, we enabled funnel logging where logging events are uploaded to the server without sending user identifiers. One of the challenges we are facing here is ensuring integrity of the logged data without relying upon user identifiers. To address this challenge we are working on adding de-identified authentication.

# Application to Problems and Results

**Label Balancing:** Here we show the impact of label balancing on model score distribution for a case of a binary classifier. For this particular problem there is (usually) one sample per/device. Initially, we approached this problem by computing the label ratio on the server side. At the serving time, we would actively drop samples from a more populous class to maintain the desired ratio. While this computation was dynamic it did not account for uncertainties that could arise during the training process itself (e.g. device drop out due to network issues or battery drain). Therefore, we needed to adopt an approach where we would depend on the label ratio computed during the training process. As shown in Figure 3, the figure on the left shows the score distribution after server side only balancing is performed. The figure on the right shows the score distribution status after the federated analytics based approach was used to balance the labels. It is evident that in order to get good model performance for this type of problem we need to rely upon the statistics obtained via federated analytics.

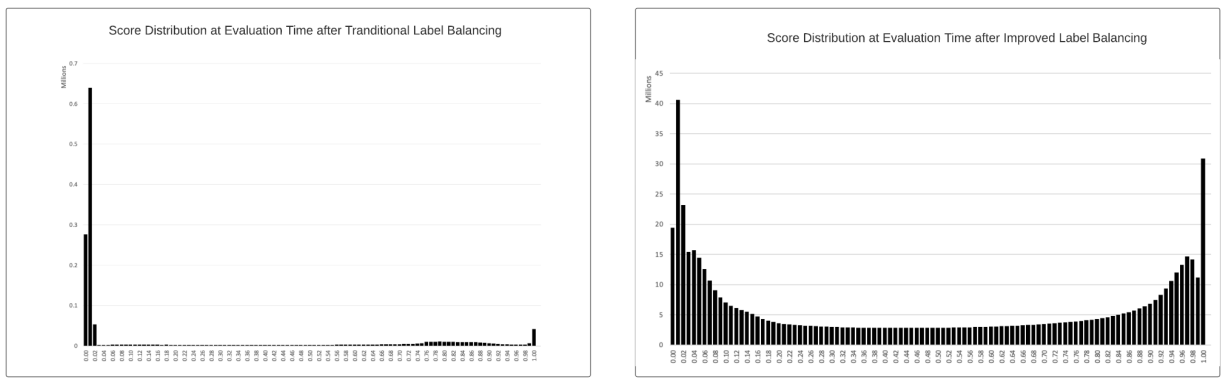

**Figure 3: Impact of Label Balancing to Score Distribution** After applying label balancing score distribution becomes more spread and not skewed towards high and low values. High performing server-side models usually do not have any skew. More uniform spread is desirable for downstream applications, such as setting a cutoff threshold. We were unable to achieve this result through alternative means (e.g. activation function selection).

**Feature Normalization**: As illustrated in Figure 4, without applying feature normalization for device only features we would face a problem where loss would saturate in the middle of training and additional rounds would have no significant effect. Similarly, accuracy would not reach a desired level. After ensuring that features are normalized using globally learned values we observed a better convergence of loss function and significant model accuracy gains.

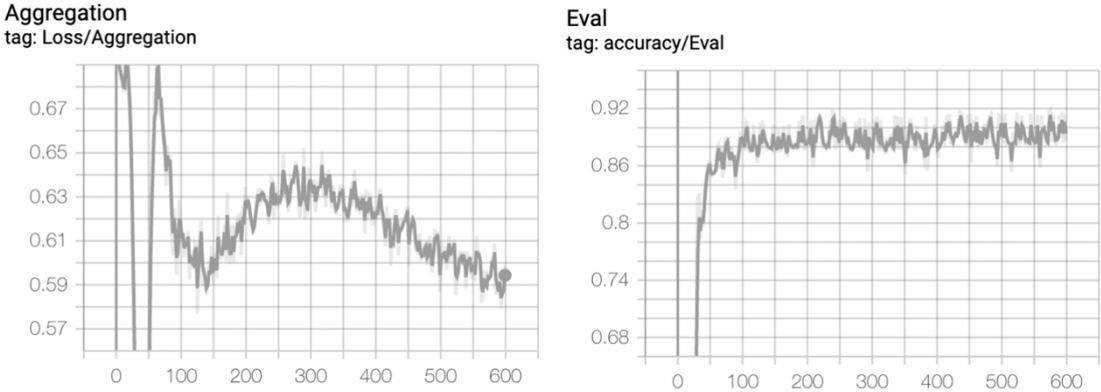

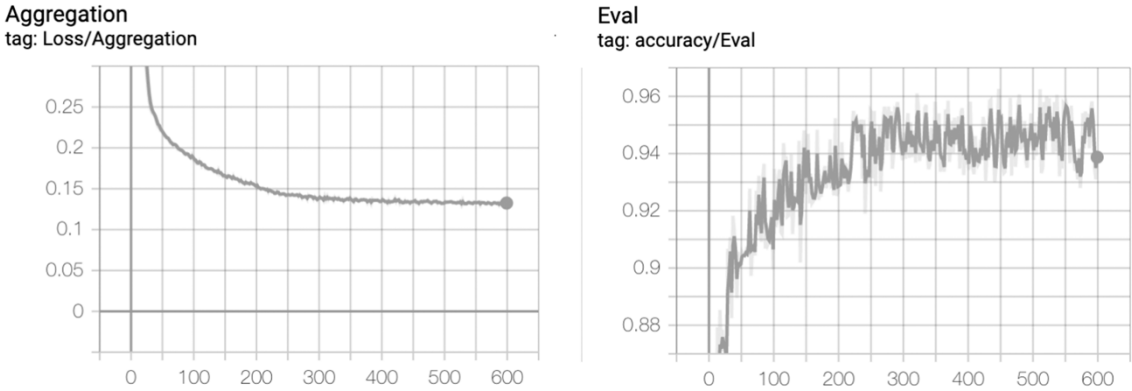

**Figure 4: Effects of Feature Normalization on Loss/Accuracy** For a binary classifier case, we observed 75% training loss reduction. Moreover, we observed about 6% average accuracy gain.

# Conclusion

This paper presents an architecture to address several challenges unique to productionizing federated machine learning with differential privacy. Those challenges include label balancing, slow release cycles, low device participation rate, privacy preserving system logging, model metric calculation and feature normalization. This paper concluded with results demonstrating the effectiveness of the proposed architecture. While this architecture is capable of successfully training and potentially deploying production federated learning models, there are several challenges left to future work. Specifically, developer speed remains one of the largest barriers to scaling production-grade federated machine learning. Current iterations of model development are several orders of magnitude slower when compared to similar sized undertakings within a centralized environment.

# Acknowledgements

We kindly thank Dzmitry Huba, Vlad Grytsun, Kaikai Wang, Aleksandar Ilic, Ilya Miranov, Ananth Raghunathan, Herb David, Kristy Lee, Curt Collins, and Komal Mangtani for substantial guidance and feedback.

# References

1. Hard, Andrew, Kanishka Rao, Rajiv Mathews, Swaroop Ramaswamy, Françoise Beaufays, Sean Augenstein, Hubert Eichner, Chloé Kiddon, and Daniel Ramage. *"Federated learning for mobile keyboard prediction."* arXiv preprint arXiv:1811.03604 (2018).
2. Xu, Jie, Benjamin S. Glicksberg, Chang Su, Peter Walker, Jiang Bian, and Fei Wang. *"Federated learning for healthcare informatics."* Journal of Healthcare Informatics Research 5, no. 1 (2021): 1-19.
3. Shaheen, Momina, Muhammad Shoaib Farooq, Tariq Umer, and Byung-Seo Kim. "*Applications of Federated Learning; Taxonomy, Challenges, and Research Trends.*" Electronics 11, no. 4 (2022): 670.
4. Cormode, Graham, and Igor L. Markov. *"Bit-efficient Numerical Aggregation and Stronger Privacy for Trust in Federated Analytics."* arXiv preprint arXiv:2108.01521 (2021).
5. Huba, Dzmitry, John Nguyen, Kshitiz Malik, Ruiyu Zhu, Mike Rabbat, Ashkan Yousefpour, Carole-Jean Wu et al. *"Papaya: Practical, Private, and Scalable Federated Learning."* arXiv preprint arXiv:2111.04877 (2021).
6. Martín Abadi, Andy Chu, Ian Goodfellow, H Brendan McMahan, Ilya Mironov, Kunal Talwar, and Li Zhang. Deep learning with differential privacy. arXiv preprint arXiv:1607.00133, 2016.
7. Pierre Stock, Igor Shilov, Ilya Mironov, Alexandre Sablayrolles. Defending against Reconstruction Attacks with Rényi Differential Privacy. arXiv preprint arXiv:2202.07623